\documentclass[letterpaper, 10 pt, journal, twoside]{IEEEtran}

\usepackage{amsmath,amsfonts}
\usepackage{array}
\usepackage{textcomp}
\usepackage{url}
\usepackage{verbatim}
\usepackage{graphicx}
\usepackage{cite}
\usepackage{makecell}
\usepackage{arydshln} 
\usepackage{multirow} 
\usepackage{gensymb}
\usepackage{booktabs} 
\usepackage{subfigure}
\usepackage{algpseudocode}
\usepackage{algorithmicx, algorithm}
\usepackage[colorlinks, citecolor=blue]{hyperref}
\hypersetup{%
	colorlinks=true,
	linkcolor=blue,
	urlcolor=blue,
	citecolor=blue,
	linkbordercolor={0 0 1}
}

\usepackage{balance} 
\newcommand{\etal}{\textit{et al.}}

\begin{document}
\title{RANSAC Back to SOTA: A Two-stage Consensus Filtering for Real-time 3D Registration}

\author{
	Pengcheng Shi, Shaocheng, Yan, Yilin Xiao, Xinyi Liu, Yongjun Zhang, Jiayuan Li 
\thanks{Manuscript received: July, 2, 2024; Revised September, 8, 2024; Accepted November, 9, 2024. This paper was recommended for publication by Editor Ashis Banerjee upon evaluation of the Associate Editor and Reviewers’ comments. This work was supported by the National Natural Science Foundation of China (NSFC) under Grant 42271444 and Grant 42201474, and the Wuhan university-Huawei Geoinformatics Innovation Laboratory Open Fund under Grant TC20210901025-2023-06. (Corresponding author: Yongjun Zhang, Jiayuan Li)}%
\thanks{Pengcheng Shi is with the School of Computer Science, Wuhan University, Wuhan 430072, China. (email: shipc$\_$2021@whu.edu.cn)}
\thanks{Yilin Xiao is with the Department of Computing, Hong Kong Polytechnic University, Hong Kong, 999077, China (email: yilin.xiao@connect.polyu.hk)}
\thanks{Shaocheng Yan, Xinyi Liu, Yongjun Zhang and Jiayuan Li are with the School of Remote Sensing and Information Engineering, Wuhan University, Wuhan 430072, China. (email: shaochengyan@whu.edu.cn, liuxy0319@whu.edu.cn, zhangyj@whu.edu.cn, ljy$\_$whu$\_$2012@whu.edu.cn)}
\thanks{Digital Object Identifier (DOI): see top of this page.}
}
\markboth{IEEE ROBOTICS AND AUTOMATION LETTERS. PREPRINT VERSION. ACCEPTED NOVEMBER, 2024}%
{SHI \MakeLowercase{\textit{et al.}}: RANSAC Back to SOTA: A Two-stage Consensus Filtering for Real-time 3D Registration}


\maketitle
\begin{abstract}
Correspondence-based point cloud registration (PCR) plays a key role in robotics and computer vision. However, challenges like sensor noises, object occlusions, and descriptor limitations inevitably result in numerous outliers. RANSAC family is the most popular outlier removal solution. However, the requisite iterations escalate exponentially with the outlier ratio, rendering it far inferior to existing methods (SC2PCR\cite{2022-CVPR-SC2PCR}, MAC\cite{2023-CVPR-MAC}, etc.) in terms of accuracy or speed. Thus, we propose a two-stage consensus filtering (TCF) that elevates RANSAC to state-of-the-art (SOTA) speed and accuracy. Firstly, one-point RANSAC obtains a consensus set based on length consistency. Subsequently, two-point RANSAC refines the set via angle consistency. Then, three-point RANSAC computes a coarse pose and removes outliers based on transformed correspondence's distances. Drawing on optimizations from one-point and two-point RANSAC, three-point RANSAC requires only a few iterations. Eventually, an iterative reweighted least squares (IRLS) is applied to yield the optimal pose. Experiments on the large-scale KITTI and ETH datasets demonstrate our method achieves up to three-orders-of-magnitude speedup compared to MAC while maintaining registration accuracy and recall. Our code is available at https://github.com/ShiPC-AI/TCF.
\end{abstract}
\section{Introduction}
\IEEEPARstart{P}{oint} cloud registration (PCR) seeks to estimate a six-degree-of-freedom (6-DOF) pose to align point clouds, which is crucial for 3D reconstruction\cite{2017-CVPR-3DMatch, 2016-ECCV-FGR, 2024-RAL-CCAG, 2020-IJGI-ANIS}, robotic navigation\cite{2023-JAG-100FPS,2020-Sensors-ANLC,2023-ISPRSJ-OPD,2024-TIV-ANHE}, and aerial photogrammetry\cite{2022-CVPR-Regtr,2024-TPR-Survey-SLAM,2023-Survey-arXiv-LPRF}. Iterative closest point (ICP)\cite{1992-TPAMI-ICP} pioneers a local registration pipeline that iteratively searches correspondences and minimizes their distances. Unfortunately, it often converges to erroneous local optima under notable viewpoint changes\cite{2021-AGCS-LIFM,shi2024indoor}. To overcome this, researchers derive coarse poses from feature correspondences\cite{2021-TRO-Teaser,2020-TGRS-CG-SAC}, known as correspondence-based PCR. However, sensor noises, object occlusions, and descriptor limitations inevitably lead to outliers, presenting great challenges for registration\cite{2019-ICCV-DCP,2023-TPAMI-QGORE}. 

Random sample consensus (RANSAC) \cite{1981-CACM-RANSAC,2009-ICRA-FPFH,2018-CVPR-GC-RANSAC} is the most popular solution to handle correspondences with outliers. It follows a \textit{generation-and-selection} strategy that iteratively samples correspondences to generate and verify hypotheses. However, the requisite iterations escalate exponentially with the outlier ratio, which renders them far inferior to existing methods in accuracy and efficiency. 
\begin{figure}[!t]
	\centering
	\includegraphics[width = 8.3cm]{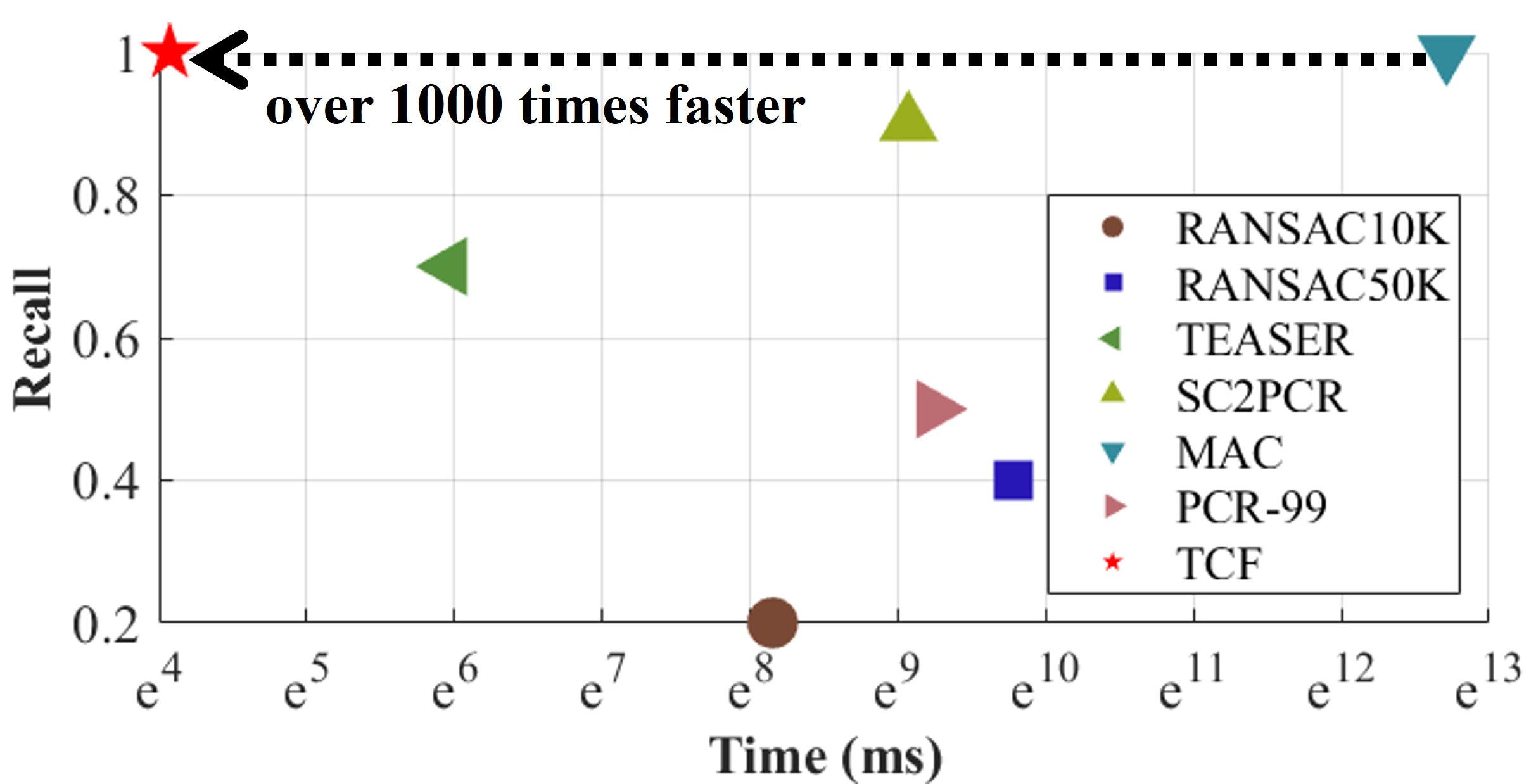}
	\caption{Average registration recall and runtime for ETH's tree sequence. The runtime denotes the average time for a single registration. Both our method and MAC achieve a 100\% recall. Remarkably, ours accomplishes this in just 59 ms, leading to a three-orders-of-magnitude improvement over MAC's 331587 ms.
	}\label{fig:recall_vs_time_trees}
\end{figure}

Thus, we propose a two-stage consensus filtering (TCF) incorporating one-point and two-point RANSAC steps before the three-point RANSAC to ensure accuracy and significantly boost efficiency. It begins with a one-point RANSAC to remove correspondence outliers via length constraints. Afterward, a two-point RANSAC evaluates the angle consistency to get a more reliable consensus. This design significantly reduces the iteration number by lowering the sampling dimension and also decreases the outlier ratio. As a result, the three-point RANSAC requires only a few iterations, enhancing registration performance. Following this, three-point RANSAC computes a coarse pose and removes outliers based on transformed correspondence's distances. Finally, a scale-adaptive Cauchy iteratively reweighted least squares (IRLS) is applied to calculate the optimal pose. As depicted in Fig. \ref{fig:recall_vs_time_trees}, our method demonstrates SOTA performance and achieves a speedup of up to three orders of magnitude. Our contributions are as follows:
\begin{itemize}	
	\item We devise a two-stage filtering method that swiftly eliminates the majority of correspondence outliers, facilitating RANSAC to reach SOTA levels.
	\item Based on our filtering method, we formulate a complete PCR pipeline. On large-scale outdoor datasets, it markedly enhances efficiency while ensuring accuracy.
\end{itemize}

\section{Related Work}\label{sec:related-work}
\subsection{Classical PCR}
\textbf{RANSAC Family}. This type of method follows a generation-and-selection strategy. GC-RANSAC\cite{2018-CVPR-GC-RANSAC} devises a locally optimized pipeline alternating between graph cut and model re-fitting. CG-SAC\cite{2020-TGRS-CG-SAC} evaluates correspondence compatibility by calculating distances between salient points and employs a compatibility-guided strategy to reduce sampling randomness. SC2-PCR\cite{2022-CVPR-SC2PCR} evaluates correspondence similarity using second-order spatial compatibility, applies a global spectral technique to identify reliable seeds, and uses a two-stage approach to expand each seed into a consensus set. To expedite processing, PCR-99\cite{2024-arXiv-PCR-99} devises an improved sample ordering guided by pairwise scale consistency and a triplet scale consistency-based outlier rejection scheme. Their performance depends on model parameter selection and requires tuning for specific problems. Runtime increases exponentially with large-scale data, especially when the outliers prevail.

\textbf{Graph Theory}. These methods model the correspondence set as a graph and use graph theories to achieve robust registration. MAC~\cite{2023-CVPR-MAC} loosens the maximum clique constraint to mine more local consensus information from a graph. It identifies maximal cliques and performs node-guided selection based on the clique with the highest graph weight. CLIPPER~\cite{lusk2021clipper} ensures dense subgraph solutions with continuous relaxation and achieves efficiency using projected gradient ascent and backtracking line search. ROBIN~\cite{shi2021robin} checks the compatibility of measurement subsets using invariants and prunes outliers by finding the maximum clique or k-core in the graph induced. TEASER~\cite{2021-TRO-Teaser} reformulates the registration problem using a truncated least squares (TLS) cost and introduces a graph-theoretic framework to decouple and sequentially solve scale, rotation, and translation estimation. Graph-based methods effectively leverage topology information and handle complex scenes but struggle with high algorithmic complexity.

\textbf{Optimization-based.} This category of methods progressively optimizes error functions to refine the pose. Iterative closest point (ICP)\cite{1992-TPAMI-ICP} pioneers a local registration pipeline that iteratively searches correspondences and minimizes Euclidean distances. Following this, several works seek to advance ICP from diverse perspectives, such as incorporating the point-to-plane metric\cite{1992-IVC-Point-Plane-ICP}, probabilistic models\cite{2019-RSS-GICP}, and branch-and-bound (BnB) theories\cite{2016-TPAMI-GOICP}. Despite robust results, poor initial guess tends to yield wrong results. Fast global registration (FGR)\cite{2016-ECCV-FGR} formulates a dense objective featuring a scaled Geman-McClure estimator as the penalty function and alleviates the impact of local minima via graduated non-convexity. Yang \etal~\cite{yang2020graduated} propose a versatile approach for robust estimation that utilizes modern non-minimal solvers, extending the applicability of Black-Rangarajan duality and graduated non-convexity to various spatial perception tasks.

\subsection{Learning-based PCR}
\textbf{End-to-end Direct Registration.} End-to-end methods optimize all steps within a unified framework, directly producing registration results through the network. Deep Closest Point (DCP)\cite{2019-ICCV-DCP} employs DGCNN\cite{2019-ATG-DGCNN} for correspondence identification and a pointer network \cite{2015-Nips-Pointer-Network} for soft matching, estimates the rigid transformation with a differentiable single value decomposition (SVD) layer. However, it struggles with partially overlapping point clouds. REGTR\cite{2022-CVPR-Regtr} employs a transformer network to estimate the probability of each point residing in the overlapping region, thereby supplanting explicit feature matching and RANSAC\cite{1981-CACM-RANSAC}. CCAG\cite{2024-RAL-CCAG} employs cross-attention mechanisms and depth-wise separable convolutions to capture point cloud relationships. Additionally, it introduces an adaptive graph convolution multi-layer perceptron (MLP) to augment node expressiveness. Although proficient with synthetic datasets, these methods may falter on real-world data, limiting their accuracy and practical utility.

\textbf{Descriptor Learning.} Descriptor learning mainly aims to create sparse descriptions for local 3D patches using a weight-sharing network or produce dense descriptions for the entire point cloud in a single forward pass. The pioneering work 3DMatch\cite{2017-CVPR-3DMatch} converts local 3D patches into volumetric representations and then derives descriptors via a siamese convolutional network. FCGF\cite{2019-ICCV-FCGF} adopts sparse tensor representation and employs Minkowski convolutional neural networks as the backbone to learn dense features. Several works successively boost the feature descriptiveness via N-tuple loss\cite{2018-CVPR-PPFNET}, folding-based encoder\cite{2018-ECCV-Ppf-foldnet}, and spherical CNNs\cite{2021-TPAMI-LEAD}. While they eliminate manual feature design and bolster robustness, substantial labeling, training, and optimization remain time-consuming, particularly for low-cost robot platforms.
\section{Method}\label{sec:method}
\subsection{Motivation}
\begin{figure}[!t]
	\centering
	\includegraphics[width = 6.5cm]{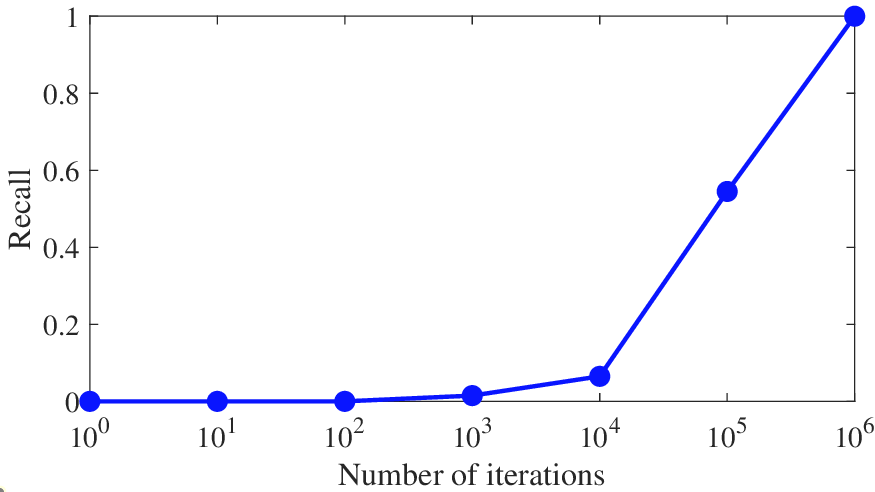}
	\caption{Registration recall of RANSAC at different iterations. We create 200 pairs of point clouds containing 2\% inliers to evaluate the registration recall.}\label{fig:recall_vs_iteration}
\end{figure}
The minimum iteration, $N_{ransac}$, required by RANSAC is computed as:
\begin{equation}
	\label{eq:ransac-nt}
	N_{ransac} = \left \lceil  \frac{\log(1-\lambda)}{\log\left(1-\left(1-\vartheta \right)^s\right)}\right \rceil
\end{equation}
where $\vartheta$ is the outlier ratio. $s$ denotes the sample size. $\left \lceil\cdot\right \rceil$ is a ceiling function. $\lambda$ represents the probability of at least one perfect minimal subset, preset to a default value of 0.99. As shown in Fig. \ref{fig:recall_vs_iteration}, we evaluate the registration recall of 200 pairs of point clouds at different iterations. As iterations increase from $10^3$ to $10^5$ and $10^6$, registration recall improves from 2\% to 5\% and eventually reaches 100\%. However, the outlier ratio in 3D correspondence is notably high, which leads to a substantial increase in RANSAC iterations based on Eq (\ref{eq:ransac-nt}). Carrying excessive iterations is impractical for real-world applications, necessitating a limit to balance efficiency and accuracy. Revisiting Eq. (\ref{eq:ransac-nt}) reveals that the sample size impacts the iteration number besides the outlier ratio. Specifically, reducing the sample size can also exponentially decrease the required iterations. Therefore, we propose a two-stage consensus filtering (TCF) comprising one-point and two-point RANSAC. One-point RANSAC applies length constraints to eliminate outliers, followed by two-point RANSAC to enhance consensus reliability via angular consistency. This approach demands fewer samples than three-point RANSAC and reduces the outlier ratio. We then feed the refined correspondence to the three-point RANSAC. With a notably decreased outlier ratio, the required iterations decrease, thus leading to enhanced performance.

\subsection{Problem Formulation}
\begin{figure}[!t]
	\centering
	\subfigure[Overall framework]{\includegraphics[width=8.4cm] {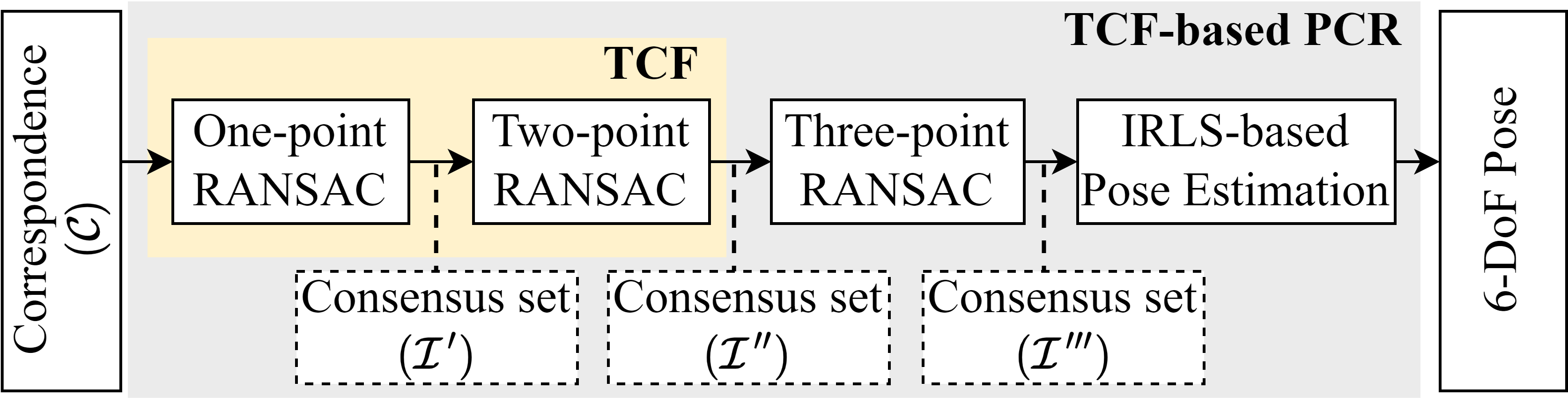}}
	\subfigure[Outlier removal]{\includegraphics[width=8.4cm] {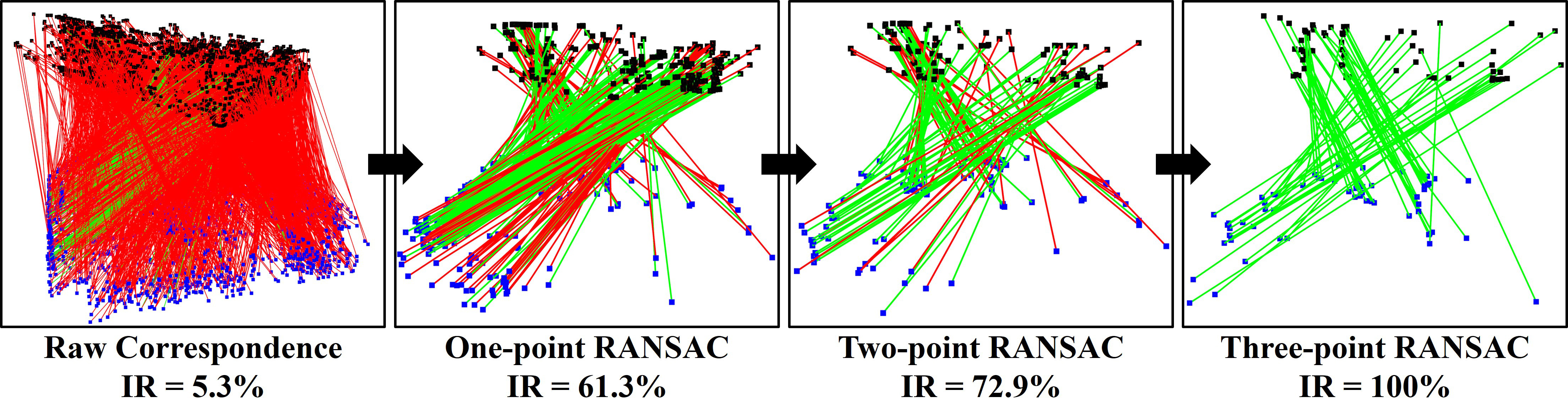}}
	\caption{Overall framework and outlier removal illustration. (a): Our method cascades one-point, two-point, and three-point RANSAC, followed by scale-adaptive Cauchy IRLS, each involving an iterative process. The raw correspondence $\mathcal{C}$ is progressively refined into subsets $\mathcal{I}^\prime$, $\mathcal{I}^{\prime\prime}$, and $\mathcal{I}^{\prime\prime\prime}$, satisfying $\mathcal{I}^{\prime\prime\prime} \subseteq \mathcal{I}^{\prime\prime} \subseteq \mathcal{I}^\prime \subseteq \mathcal{C}$. The scale-adaptive Cauchy IRLS computes the final 6-DoF pose from $\mathcal{I}^{\prime\prime\prime}$. (b): The blue and black points represent the 3D correspondences. The green lines indicate inliers and red lines represent outliers. IR stands for the inlier ratio.}\label{fig:pipeline}
\end{figure}

Given two point clouds $P=\left\{p_i \in \mathbb{R}^3, i=1,...,N_p\right\}$ and $Q=\left\{q_i \in \mathbb{R}^3, i=1,...,N_q\right\}$ to be aligned, we begin by extracting local features to establish initial correspondences $\mathcal{C} = \left\{c_k = (p_k, q_k), k=1,...,N_c\right\}$. We employ a two-stage filtering process, i.e., one-point and two-point RANSAC, to obtain the maximum consensus set. Eventually, we employ three-point RANSAC and IRLS to estimate the rigid transformation: a rotation matrix $\mathbf{R}\in\mathbb{R}^{3\times3}$ and a translation vector $\boldsymbol{t}\in\mathbb{R}^3$. Fig. \ref{fig:pipeline} illustrates our registration pipeline.
\subsection{One-point RANSAC}
\begin{figure}[!t]
	\centering
	\includegraphics[width = 7.5cm]{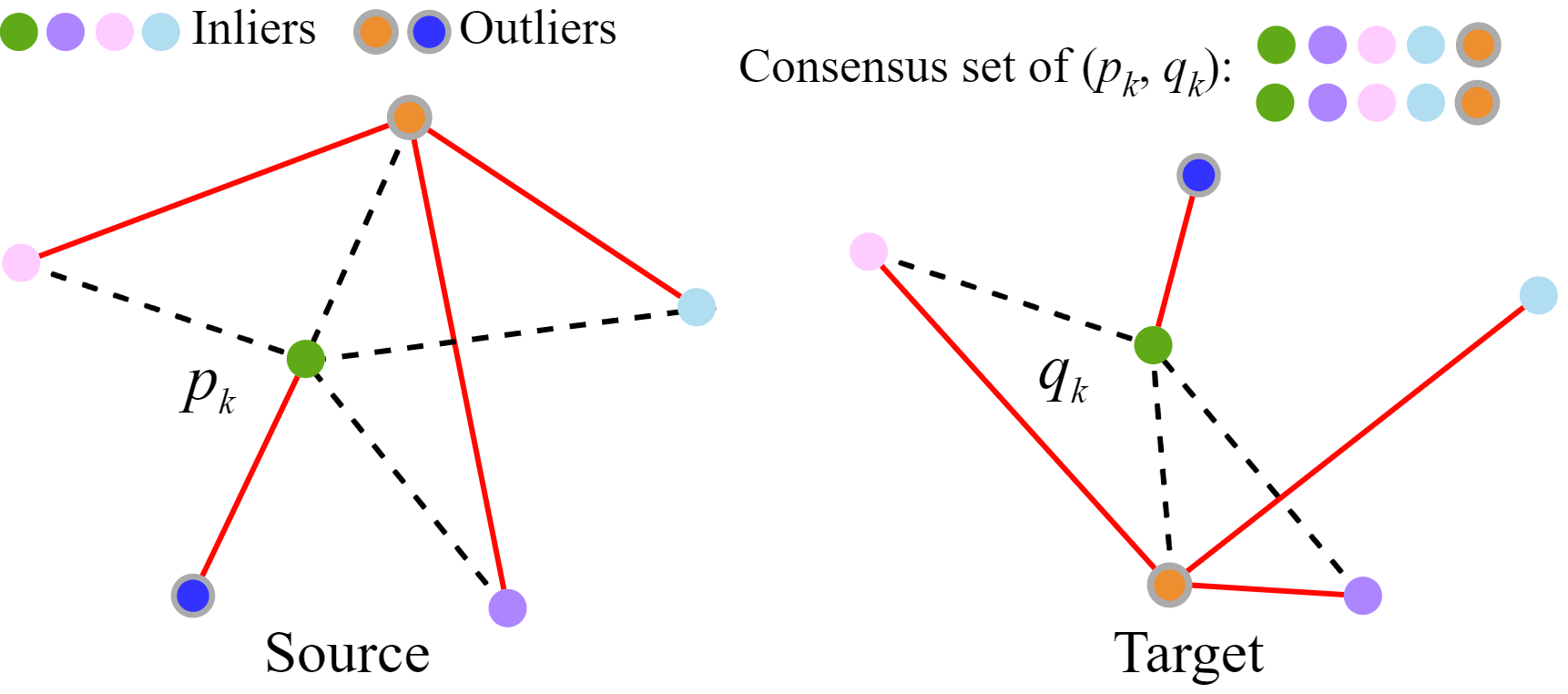}
	\caption{One-point consensus generation. Same-colored points indicate a point correspondence. Dotted black lines show correct edge correspondences, while red indicates erroneous ones. An edge correspondence is correct if their length discrepancy $< 2\tau$. For $(p_k, q_k)$, applying length consistency identifies a maximal consensus comprising four inliers and one outlier marked in yellow.}\label{fig:outlier-1pt-ransac}
\end{figure}

\begin{algorithm}[!t]
	\small 
	\caption{One-point RANSAC} %
	{\bf Input:} Correspondences $\mathcal{C}$\\
	{\bf Initialize}: $\lambda = 0.99$, $|\mathcal{I}^\prime|=0, i = 0$, $	N_{ransac}^{(1)}=1$. 
	\begin{algorithmic}[1]
		\While {$i \le N_{ransac}^{(1)}$}
		\State Randomly select a point correspondence $c_k$ from $\mathcal{C}$
		\State Find a consensus set $\mathcal{I}_i$ including $c_k$ via Eq. (\ref{eq:consensus-search-1}) and (\ref{eq:length-consistency})
		\If{$|\mathcal{I}_i| \ge |\mathcal{I}^\prime|$}
		\State $|\mathcal{I}^\prime|= |\mathcal{I}_i|, \mathcal{I}^\prime = \mathcal{I}_i$
		\State Update $	N_{ransac}^{(1)}$ using Eq. (\ref{eq:nt-1pt})
		\EndIf
		\State $i=i+1$
		\EndWhile
	\end{algorithmic}
	{\bf Output:} Correspondences $\mathcal{I}^\prime$
	\label{algo:1pt-ransac}
\end{algorithm}
As described in Algorithm \ref{algo:1pt-ransac}, one-point RANSAC takes the initial correspondence $\mathcal{C}$ as input and outputs a consensus set $\mathcal{I}^\prime$ based on length consistency. The main steps are as follows:

\textbf{One-point Consensus Generation.} As depicted in Fig. \ref{fig:outlier-1pt-ransac}, for a random point correspondence $c_k=(p_k, q_k)$, we compare it with others to gather a maximal consensus set $\mathcal{I}_i$.
\begin{equation}
	\label{eq:consensus-search-1}
	\mathcal{I}_i = \{ c_j \, | \, c_j \in \mathcal{C}, \forall j, d(c_j, c_k) < 2\tau \}
\end{equation}
\begin{equation}
	\label{eq:length-consistency}
	d(c_j, c_k) = \left|\left|\left|p_j-p_k\right|\right|_2 - \left|\left|q_j-q_k\right|\right|_2\right|
\end{equation}
where $\mathcal{I}_i \subseteq \mathcal{C}$ is a correspondence subset including $c_k$. $i$ denotes the current iteration number.  $||\cdot||_2$ means the $L_2$ norm. $k$ and $j$ are two indices of point correspondences. If $c_j$ and $c_k$ are inliers, they should satisfy $d(c_j, c_k) < 2\tau$. $\tau$ is a noise bound. Our method automatically adjusts the noise bounds as the noise level increases, ensuring the inliers continue to meet the criteria. Thus, it remains effective despite higher noise.

\textbf{Consensus Verification.} Upon identifying a larger consensus during the iteration, we update the minimal iteration requirement  $N_{ransac}^{(1)}$.
\begin{equation}
	\label{eq:nt-1pt}
		N_{ransac}^{(1)} = \left \lceil  \frac{\log(1-\lambda)}{\log\left(1-\frac{|\mathcal{I}^\prime|}{|\mathcal{C}|}\right)}\right \rceil
\end{equation}
where superscript (1) indicates one-point sampling. $\mathcal{I}^\prime$ denotes the current optimal consensus set. When the iteration number $i$ surpasses $N_{ransac}^{(1)}$, the algorithm terminates and outputs $\mathcal{I}^\prime$.
\subsection{Two-point RANSAC}
While the consensus set meets length requirements, outliers like the yellow correspondence still exist. Thus, we propose a two-point RANSAC to refine the consensus. As summarized in Algorithm \ref{algo:2pt-RANSAC}, two-point RANSAC takes the result $\mathcal{I}^\prime$ of one-point RANSAC as input and excludes outliers based on angle consistency. The main steps are as follows:
\begin{algorithm}[!t]
	\small 
	\caption{Two-point RANSAC} %
	{\bf Input:} Correspondences $\mathcal{I}^\prime$\\
	{\bf Initialize}: $\lambda = 0.99$, $|\mathcal{I}^{\prime\prime}|=0, i^\prime = 0$, $	N_{ransac}^{(2)}=1$. 
	\begin{algorithmic}[1]
		\While {$i^\prime \le N_{ransac}^{(2)}$}
		\State Randomly select two point correspondences $c_i, c_j \in \mathcal{I}^\prime$
		\State Find a subset $\mathcal{S}_{ij} \subseteq \mathcal{I}^\prime$ using Eq. (\ref{eq:subset-2pt-ransac})
		\State Find a consensus set $\mathcal{I}_{i^\prime} \subseteq \mathcal{S}_{ij}$ using Eq. (\ref{eq:consensus-2pt})-(\ref{eq:beta-2pt}) 			
		\If{$|\mathcal{I}_{i^\prime}| \ge |\mathcal{I}^{\prime\prime}|$}
		\State $|\mathcal{I}^{\prime\prime}|= |\mathcal{I}_{i^\prime}|, \mathcal{I}^{\prime\prime} = \mathcal{I}_{i^\prime}$
		\State Update $	N_{ransac}^{(2)}$ based on Eq. (\ref{eq:nt-2pt})	    
		\EndIf
		\State $i^\prime = i^\prime + 1$		
		\EndWhile
	\end{algorithmic}
	{\bf Output:} Correspondences $\mathcal{I}^{\prime\prime}$
	\label{algo:2pt-RANSAC}
\end{algorithm}

\textbf{Two-point Subset Sampling.} We begin by randomly selecting two different point correspondences, $c_i=(p_i, q_i)$ and $c_j=(p_j, q_j)$ from $\mathcal{I}^\prime$. Then, we compare them with others to collect a subset $\mathcal{S}_{ij}$ that simultaneously satisfies two  length consistencies:
\begin{equation}
	\label{eq:subset-2pt-ransac}
	\mathcal{S}_{ij}= \{ c_k \, | \, c_k \in \mathcal{I}^\prime, \forall k, d(c_k, c_i) < 2\tau, d(c_k, c_j) < 2\tau \}
\end{equation}
where $c_i, c_j \in \mathcal{S}_{ij}$ is a correspondence subset of $\mathcal{I}^{\prime}$. $d(\cdot)$ is the length discrepancy in Eq. (\ref{eq:length-consistency}).

\begin{figure}[!t]
	\centering
	\includegraphics[width = 7.5cm]{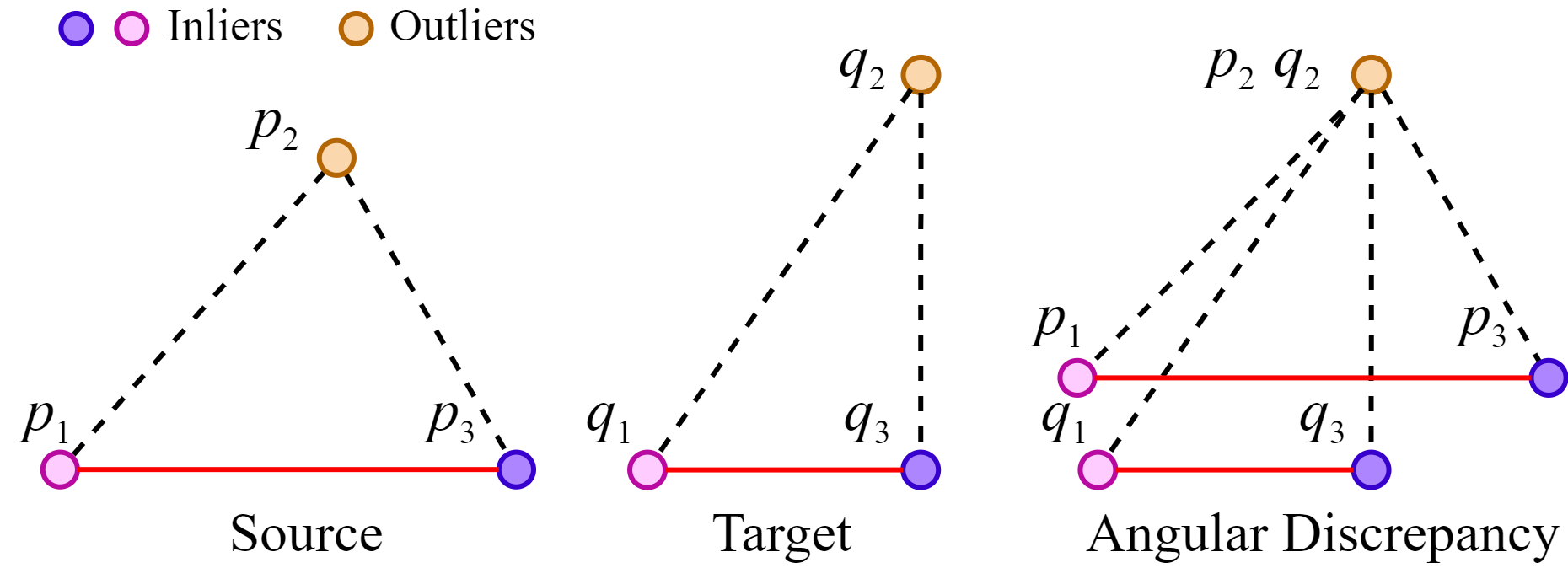}
	\caption{Angular discrepancy. The circled letter signifies a point, and the same color denotes a point correspondence. Black dotted lines represent correct edge correspondences, i.e., $|p_1p_2-q_1q_2| < 2\tau$ and $|p_3p_2-q_3q_2|< 2\tau$. The yellow outlier produces triangles with two similar edge lengths but different shapes due to angle discrepancies.}\label{fig:outlier-2pt-ransac}
\end{figure}

\begin{figure}[!t]
	\centering
	\includegraphics[width = 7.5cm]{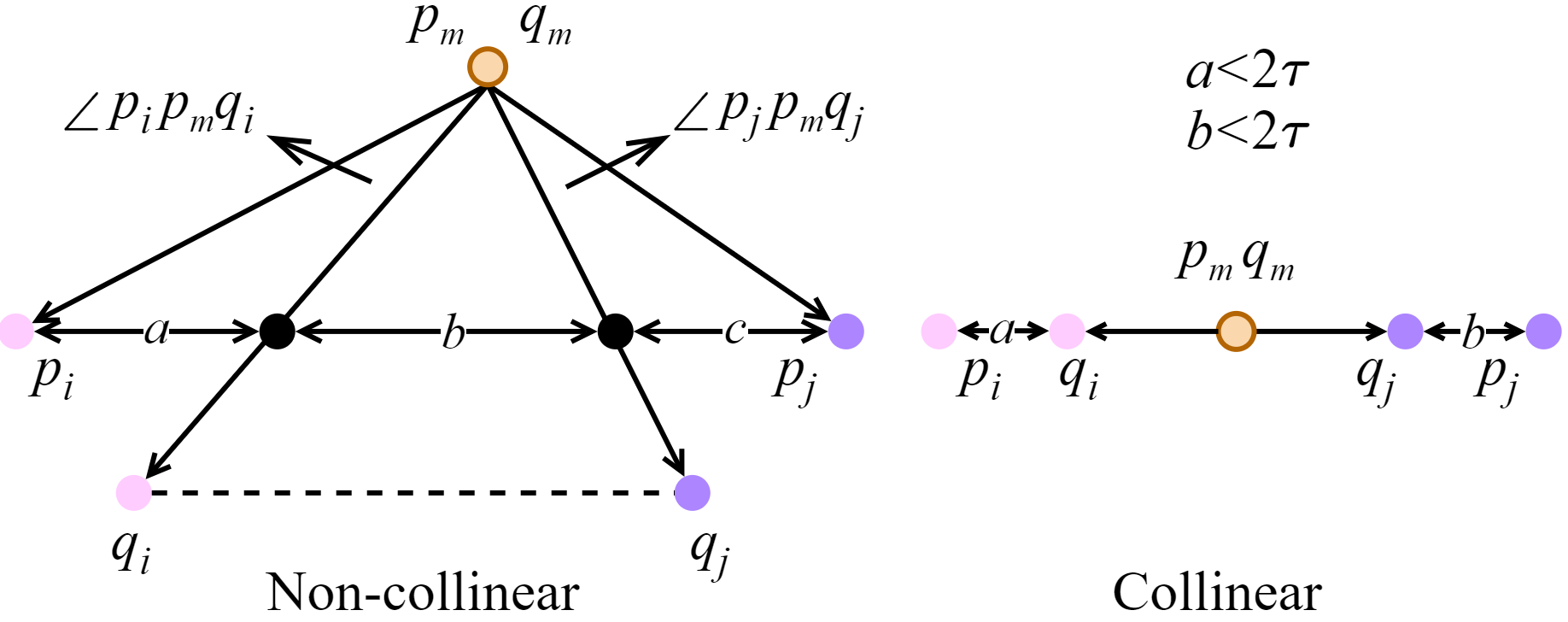}
	\caption{$\beta$ and two cases of correspondence distribution. We overlap the correspondence $c_m=(p_m, q_m)$ for clarity. Given that $c_i$ and $c_j$ are two inliers, they satisfy $\left|\left|p_i p_j\right| - \left|q_i q_j\right|\right| \leq 2\tau$. If $c_m$ is also an inlier, then $\left| \left|p_mp_i\right|-\left|q_mq_i\right|\right| < 2\tau$ and $\left| \left|p_mp_j\right|-\left|q_mq_j\right|\right| < 2\tau$. To ensure robust consensus, we further constrain the angle discrepancy $\alpha(c_m, c_i, c_j) = (\angle{p_ip_mq_i} + \angle{p_jp_mq_j}) < \beta$, where $\angle{p_ip_mq_i} \approx \arcsin\frac{a}{\left| p_mp_i\right|}$, $\angle{p_jp_mq_j} \approx \arcsin\frac{c}{\left| p_mp_j\right|}$, and $a, c < 2\tau$. We set $\beta$ to $\left|\arcsin\frac{\tau}{|p_mp_i| }+\arcsin\frac{\tau}{|p_mp_j|}\right|$.}\label{fig:beta-1-2}
\end{figure}

\textbf{Consensus Generation} As illustrated in Fig. \ref{fig:outlier-2pt-ransac}, despite two edge correspondences maintaining length consistency, two triangles still exhibit distinct shapes due to angle discrepancy. Therefore, we propose angle consistency to encourage point correspondences to form roughly congruent triangles, yielding a more robust consensus set $\mathcal{I}_{i^\prime}$.  
\begin{equation}
	\label{eq:consensus-2pt}
	\mathcal{I}_{i^\prime} = \mathop{{\rm argmax}}\limits_{\mathcal{I} \subseteq \mathcal{S}_{ij}} |\mathcal{I}|, \forall \, c_m \in \mathcal{I}, \alpha(c_m, c_i, c_j) < \beta
\end{equation}
\begin{equation}
	\label{eq:alpha-angle}
	\alpha(c_m, c_i, c_j) = \angle{p_ip_mq_i} + \angle{p_jp_mq_j}
\end{equation}
\begin{equation}
	\label{eq:beta-2pt}
	\beta = \left|\arcsin\frac{\tau}{|p_mp_i| }+\arcsin\frac{\tau}{|p_mp_j|}\right|
\end{equation}
where $i^\prime$ means the current iteration number. $\alpha(c_m, c_i, c_j)$ denotes the angle disparity. Fig. \ref{fig:beta-1-2} depicts the computation of $\beta$ and two cases of correspondence distribution.

\textbf{Consensus Verification.} This step is similar to that of the one-point RANSAC. Upon identifying a larger inlier set during the iteration, we immediately update the iteration's termination condition $N_{ransac}^{(2)}$.
\begin{equation}
	\label{eq:nt-2pt}
	N_{ransac}^{(2)} =\left \lceil \frac{\log(1-\lambda)}{\log\left(1-\left(\frac{|\mathcal{I}^{\prime\prime} |}{|\mathcal{I}^\prime|}\right)^2\right)}\right \rceil 
\end{equation}
This step is similar to that of the one-point RANSAC. Upon identifying a larger inlier set during the iteration, we immediately update the iteration's termination condition $N_{ransac}^{(2)}$.
\subsection{Three-point RANSAC}
After one-point and two-point RANSAC, we obtain a correspondence subset satisfying length and angle constraints. Despite a notable increase in the inlier ratio, outliers may remain. Hence, we use three-point RANSAC to identify a consensus set $\mathcal{I}^{\prime\prime\prime}$ based on the distances of transformed correspondences.
\begin{equation}
	\label{eq:trans-estimate-ransac-3}
	\begin{aligned}
		 \mathcal{I}^{\prime\prime\prime} = &\mathop{{\rm argmax}}\limits_{\mathbf{R}_k, \boldsymbol{t}_k} |\mathcal{I}|, \mathcal{I} \subseteq \mathcal{I}^{\prime\prime}\\ 
			&s.t., \forall (p_i, q_i) \in \mathcal{I}, |\mathbf{R}_kp_i+\boldsymbol{t}_k-q_i| < \tau.
	\end{aligned}
\end{equation}
where $k$ means the iteration number. $\mathcal{I}$ represents a correspondence subset, and $(p_i, q_i)$ denotes a point correspondence. We employ singular value decomposition (SVD) to compute the rotation matrix $\mathbf{R}_k$ and translation vector $\boldsymbol{t}_k$ from three non-collinear correspondences.

\subsection{Scale-adaptive Cauchy IRLS}
\begin{algorithm}[!t]
	\small 
	\caption{Scale-adaptive Cauchy IRLS} %
	{\bf Input:} Correspondences $\mathcal{I}^{\prime\prime\prime}$\\
	{\bf Initialize}: $j=1$, $N_j = 100$, $w_1=\{1\}_{1:|\mathcal{I}^{\prime\prime\prime}|}$, $|\mathcal{I}_1| = |\mathcal{I}^{\prime\prime\prime}|, \mu = 1.3, e_{min}=0.01, \gamma_{min}=1.0$
	\begin{algorithmic}[1]
		\State $\gamma_1$ $\xleftarrow{\max}$ residuals $e$ $\xleftarrow{\mathbf{R}, \boldsymbol{t}} {\rm SVD}(\hat{\mathcal{I}}^\prime_1)$
		\While {$j \le N_j$}
		\State Estimate residuals $e_j$ $\xleftarrow{\mathbf{R}_j, \boldsymbol{t}_j} {\rm IRLS}(w_j, \mathcal{I}_j)$
		\State Update inlier set $\mathcal{I}_{j+1}$ by $e_j < 3 \gamma_j$
		\State Update inlier residuals $e_{j+1}^\prime = e_j[\mathcal{I}_{j+1}] \subseteq e_j$ 
		\State Update inlier weights $w_{j+1} = \frac{\gamma_j^2}{\gamma_j^2 + e_{j+1}^\prime}$ 
		\State Update scale factor $\gamma_{j+1} = \gamma_{j} / \mu$ 
		\If{$|w_{j+1}*{e_{j+1}^\prime}^2 - w_j*{e_j^\prime}^2|$ $\textless$ $e_{min}$ or $\gamma_{j+1}$ $\textless$ $\gamma_{min}$}
		\State $\mathbf{R}^* = \mathbf{R}_j, \boldsymbol{t}^* = \boldsymbol{t}_j$
		\State \textbf{break}
		\EndIf
		\State $j=j+1$
		\EndWhile
	\end{algorithmic}
	{\bf Output:} Transformation $(\mathbf{R}^*, \boldsymbol{t}^*)$
	\label{algo:SA-Cauchy-IRLS}
\end{algorithm}
Traditional Cauchy estimation exhibits considerable deviation from the ground truth in the first iteration if the outliers exceed 50$\%$. Correct inliers incur higher residuals and receive smaller weights. Thus, we introduce a scale-adaptive Cauchy IRLS\cite{2023-AGCS-SACR} in Algorithm \ref{algo:SA-Cauchy-IRLS}. We start by assigning equal weights ($w$) to all correspondences ($\mathcal{I}^{\prime\prime\prime}$). We then progressively lower the residual threshold based on a scale factor ($\gamma$) and preserve inliers ($\mathcal{I}$). $\gamma$ decreases continuously via the constant $\mu=1.3$. Upon iteration termination, we output the optimal transformation ($\mathbf{R}^*, \boldsymbol{t}^*$). Unlike FGR\cite{2016-ECCV-FGR}, scale-adaptive Cauchy IRLS uses a novel Cauchy objective function and automatically excludes outliers with control parameters during optimization. Additionally, it provides the applications of space intersection, robust feature matching, and point cloud registration.
\section{Experiments}\label{sec:experiments}
\begin{table}[!t]
	\renewcommand{\arraystretch}{1}
	\centering
	\caption{Dataset details. $D_p$ means the average pairwise distance. $|\mathcal{C}|$ refers to the average number of correspondences. $\rho$ is the average inlier ratio.}
	\begin{tabular}{c|c|ccccc}
		\hline
		\multicolumn{2}{c|}{Sequences}&Type&\makecell[c]{$D_p$\\(m)}&\makecell[c]{Pairs\\(/)}&\makecell[c]{$|\mathcal{C}|$\\(/)}&\makecell[c]{$\rho$\\(/)}\\\hline
		\multirow{5}{*}{\rotatebox{90}{ETH}}&Arch&Outdoor&16.48&8&6989&1.1\%\\
		&Courtyard&Outdoor&16.63&28&9910&6.1\%\\
		&Facade&Outdoor&4.06&21&2250&12.2\%\\
		&Office&Indoor&7.49&8&2964&5.6\%\\
		&Trees&Outdoor&10.64&10&8242&1.2\%\\\hline
		\multirow{5}{*}{\rotatebox{90}{KITTI}}&01&Highway&16.97&200&3343&2.4\%\\
		&02&Urban+Country&16.97&200&3104&3.5\%\\
		&03&Country&17.02&200&3175&4.6\%\\
		&06&Urban&16.95&200&3655&3.6\%\\
		&09&Urban+Country&16.96&200&3042&3.8\%\\
		\hline
	\end{tabular}\label{tab:dataset}
\end{table}
\subsection{Experimental Setup}
\textbf{Datasets.} As summarized in Table \ref{tab:dataset}, we select ETH\cite{2015-ISPRSJ-GCRO} and KITTI\cite{2012-Dataset-CVPR-KITTI} for validation. The ETH dataset comprises 32 scans from five scenes (office, facade, courtyard, arch, and trees), acquired using Z + F Imager 5006i and Faro Focus 3D scanners. The KITTI dataset provides 11 Velodyne HDL-64E LiDAR sequences across diverse scenarios. We choose sequences 01 (highway), 02 (urban+country), 03 (country), 06 (urban), and 09 (urban+country). We randomly sample 200 pairs of point clouds per sequence.

\textbf{Evaluation Metrics.} We evaluate all methods via registration recall $rr= \frac{N_{success}}{N_{all}}$, rotation error $e_r= \arccos\frac{tr(\mathbf{R}_e(\mathbf{R}_g)^{\rm T})-1}{2}$, and translation error $e_t=||\boldsymbol{t}_e-\boldsymbol{t}_g||$. $tr(\cdot)$ is the matrix trace. $\boldsymbol{t}$ and $\mathbf{R}$ denote translation and rotation, respectively. Subscript $g$ denotes ground-truth values while $e$ represents estimated ones. $N_{success}$ is the count of successfully registered point cloud pairs while $N_{all}$ is the total number. Registration is successful if $e_r \leq 5\degree, e_t \leq 0.5\ m$ on the ETH dataset, and $e_r \leq 5\degree, e_t \leq 1\ m$ on the KITTI dataset.

\textbf{Implementation Details.} We select RANSAC10K\cite{1981-CACM-RANSAC}, RANSAC50K\cite{1981-CACM-RANSAC}, TEASER\cite{2021-TRO-Teaser}, SC2-PCR\cite{2022-CVPR-SC2PCR}, MAC\cite{2023-CVPR-MAC}, and PCR-99\cite{2024-arXiv-PCR-99} for comparisons. We implement our method in C++ on Ubuntu 20.04, running all experiments on an Intel Core i9-10850K CPU. For ETH, we downsample each point cloud with 0.1 m voxels, then extract ISS\cite{2009-ICCVW-ISS} keypoints and FPFH\cite{2009-ICRA-FPFH} descriptors. We apply 0.3 m voxels for KITTI, randomly sample 5,000 points, and extract FPFH\cite{2009-ICRA-FPFH} descriptors. Two points qualify as a correspondence if their descriptors rank in each other's top three. We set $\tau$ as the point cloud's resolution. The maximal distance for a valid correspondence is 0.3 m in the ETH dataset and 0.6 m for the KITTI dataset.
\subsection{Results on ETH}
\begin{table*}[!t]
	\renewcommand{\arraystretch}{1}
	\centering
	\caption{Registration results on the ETH dataset. $|\mathcal{C}|$ refers to the average number of correspondences, while $\rho$ is the average inlier ratio. Bold and underline denote the first and second place, respectively. $-$ means that all trials have failed.}
	\begin{tabular}{m{1.5cm}<{\centering}|m{0.2cm}<{\centering}m{0.31cm}<{\centering}m{0.31cm}<{\centering}m{0.68cm}<{\centering}|m{0.2cm}<{\centering}m{0.31cm}<{\centering}m{0.31cm}<{\centering}m{0.68cm}<{\centering}|m{0.2cm}<{\centering}m{0.31cm}<{\centering}m{0.31cm}<{\centering}m{0.68cm}<{\centering}|m{0.2cm}<{\centering}m{0.31cm}<{\centering}m{0.31cm}<{\centering}m{0.68cm}<{\centering}|m{0.2cm}<{\centering}m{0.31cm}<{\centering}m{0.31cm}<{\centering}m{0.68cm}<{\centering}}
		\hline
		\multirow{3}{*}{Methods}&\multicolumn{4}{c|}{Arch}&\multicolumn{4}{c|}{Courtyard}&\multicolumn{4}{c|}{Facade}&\multicolumn{4}{c|}{Office}&\multicolumn{4}{c}{Trees}\\\cline{2-21}
		&$rr$&$e_r$&$e_t$&time&$rr$&$e_r$&$e_t$&time&$rr$&$e_r$&$e_t$&$t$&$rr$&$e_r$&$e_t$&time&$rr$&$e_r$&$e_t$&time\\
		&(\%)&(\degree)&(m)&(ms)&(\%)&(\degree)&(m)&(ms)&(\%)&(\degree)&(m)&(ms)&(\%)&(\degree)&(m)&(ms)&(\%)&(\degree)&(m)&(ms)\\\hline
		RANSAC10K&$-$&$-$&$-$&$-$&89&0.47&0.12&3863&\textbf{100}&0.41&0.09&1802&90&1.35&0.16&2022&20&1.11&0.20&3470\\
		RANSAC50K&38&0.72&0.28&16263&100&0.21&0.10&18981&100&0.30&0.07&9047&90&1.23&0.22&10243&40&1.46&0.21&17752\\
		TEASER&\underline{63}&0.25&0.07&\underline{318}&\textbf{100}&0.06&\textbf{0.04}&\underline{521}&\textbf{100}&0.22&\underline{0.03}&\underline{67}&90&0.70&0.07&\underline{81}&70&\textbf{0.12}&\textbf{0.03}&\underline{391}\\
		SC2PCR&\textbf{88}&0.18&0.06&6548&\textbf{100}&\underline{0.04}&\textbf{0.04}&7383&\textbf{100}&\textbf{0.08}&\textbf{0.02}&693&\textbf{100}&\textbf{0.45}&\textbf{0.04}&1204&\underline{90}&0.16&\textbf{0.03}&8699\\
		MAC&\textbf{88}&\textbf{0.09}&\underline{0.05}&316444&\textbf{100}&\underline{0.04}&\textbf{0.04}&797857&\textbf{100}&0.13&\textbf{0.02}&14881&\textbf{100}&\underline{0.50}&\underline{0.06}&63080&\textbf{100}&0.25&0.07&331587\\
		PCR-99&50&0.18&\textbf{0.04}&7718&\textbf{100}&0.05&\textbf{0.04}&13881&95&0.22&0.04&774&90&0.85&0.08&1362&50&\underline{0.15}&\underline{0.04}&10309\\
		Ours&\textbf{88}&\underline{0.14}&\textbf{0.04}&\textbf{35}&\textbf{100}&\textbf{0.02}&\textbf{0.04}&\textbf{96}&\textbf{100}&\underline{0.11}&\underline{0.03}&\textbf{35}&\textbf{100}&0.64&\underline{0.06}&\textbf{27}&\textbf{100}&0.19&\textbf{0.03}&\textbf{58}\\
		\hline
	\end{tabular}\label{tab:eva-eth-top-3}
\end{table*}
\begin{figure}[!t]
	\centering
	\includegraphics[width = 8cm]{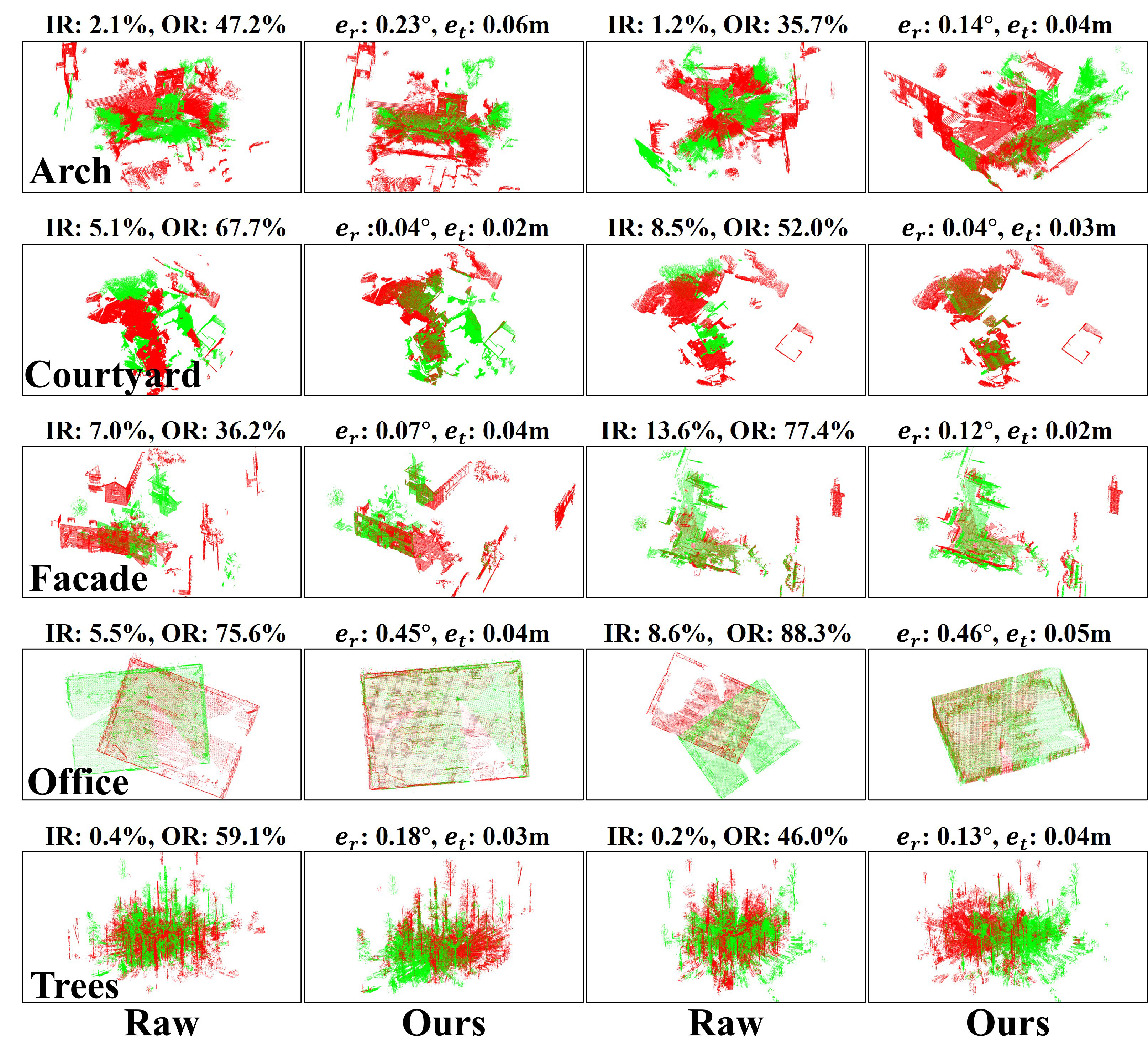}
	\caption{Visualization of registration results on the ETH dataset. Red and green denote the source and target point clouds, respectively. IR refers to the inlier ratio, and OR denotes the overlap ratio. $e_r$ refers to the rotation error, and $e_t$ denotes the translation error.}\label{fig:eth-registration}
\end{figure}
\begin{table*}[t]
	\renewcommand{\arraystretch}{1}
	\centering
	\caption{Registration results on the KITTI dataset. $|\mathcal{C}|$ refers to the average number of correspondences. $\rho$ is the average inlier ratio. Bold and underline denote the first and second place, respectively.}
	\begin{tabular}{m{1.5cm}<{\centering}|m{0.2cm}<{\centering}m{0.31cm}<{\centering}m{0.31cm}<{\centering}m{0.68cm}<{\centering}|m{0.2cm}<{\centering}m{0.31cm}<{\centering}m{0.31cm}<{\centering}m{0.68cm}<{\centering}|m{0.2cm}<{\centering}m{0.31cm}<{\centering}m{0.31cm}<{\centering}m{0.68cm}<{\centering}|m{0.2cm}<{\centering}m{0.31cm}<{\centering}m{0.31cm}<{\centering}m{0.68cm}<{\centering}|m{0.2cm}<{\centering}m{0.31cm}<{\centering}m{0.31cm}<{\centering}m{0.68cm}<{\centering}}
		\hline
		\multirow{3}{*}{Methods}&\multicolumn{4}{c|}{01}&\multicolumn{4}{c|}{02}&\multicolumn{4}{c|}{03}&\multicolumn{4}{c|}{06}&\multicolumn{4}{c}{09}\\\cline{2-21}
		&$rr$&$e_r$&$e_t$&time&$rr$&$e_r$&$e_t$&time&$rr$&$e_r$&$e_t$&$t$&$rr$&$e_r$&$e_t$&time&$rr$&$e_r$&$e_t$&time\\
		&(\%)&(\degree)&(m)&(ms)&(\%)&(\degree)&(m)&(ms)&(\%)&(\degree)&(m)&(ms)&(\%)&(\degree)&(m)&(ms)&(\%)&(\degree)&(m)&(ms)\\\hline
		RANSAC10K&42&1.10&0.42&2206&69&1.16&0.42&2073&97&1.06&0.35&2098&88&0.99&0.38&2232&85&1.25&0.36&2063\\
		RANSAC50K&44&0.74&0.30&11102&73&0.97&0.36&10268&\textbf{100}&0.78&0.28&10295&87&0.75&0.29&10854&88&1.06&0.32&9931\\
		TEASER&52&0.65&0.26&\underline{166}&\underline{74}&1.00&\underline{0.34}&\underline{117}&\underline{99}&0.86&\underline{0.25}&\underline{115}&\underline{97}&0.65&0.22&\underline{134}&89&1.05&0.28&\underline{106}\\
		SC2PCR&49&\textbf{0.36}&\textbf{0.21}&1801&\textbf{75}&\textbf{0.66}&\textbf{0.31}&1338&\textbf{100}&\textbf{0.55}&\textbf{0.23}&1387&92&\textbf{0.40}&\textbf{0.19}&1829&\underline{90}&\textbf{0.66}&\textbf{0.23}&1279\\
		MAC&\underline{54}&\underline{0.40}&0.26&35632&\underline{74}&\underline{0.72}&0.35&29158&\textbf{100}&\underline{0.56}&0.27&75199&95&\underline{0.42}&0.22&112980&\textbf{91}&\underline{0.69}&0.28&26109\\
		PCR-99&55&1.37&0.45&12219&60&1.59&0.43&5506&77&1.36&0.39&4355&65&1.45&0.43&10266&71&1.56&0.40&3078\\
		Ours&\textbf{65}&0.54&\underline{0.25}&\textbf{79}&\textbf{75}&0.73&\underline{0.34}&\textbf{58}&\textbf{100}&0.65&\underline{0.25}&\textbf{63}&\textbf{98}&0.48&\underline{0.21}&\textbf{53}&\underline{90}&0.73&\underline{0.25}&\textbf{45}\\\hline
	\end{tabular}\label{tab:eva-kitti-top-3}
\end{table*}

\textbf{Recall and Accuracy.} Fig. \ref{fig:eth-registration} depicts our registration results on the ETH dataset and Table \ref{tab:eva-eth-top-3} outlines the registration results. Courtyard, facade, and office have inlier ratios exceeding 5\% while challenging arch and trees only have around 1\%. Our method matches MAC in the recall, achieving top performance in all five scenarios. It delivers the best translation accuracy in the arch (0.04 m), courtyard (0.04 m), and trees (0.03 m), the second-best in the facade (0.03 m) and office (0.06 m).It also excels in rotational accuracy, leading in the courtyard (0.02\degree) and ranking second in the facade (0.11\degree). Our rotation error exceeds MAC, SC2PCR, and TEASER by 0.05\degree, 0.09\degree, and 0.07\degree in the arch, office, and trees, respectively. These results showcase that our method achieves comparable or superior registration accuracy and recall to mainstream methods.

\textbf{Running Efficiency.} Our method is generally two to three orders of magnitude faster than baseline methods. For example, on arch, it outperforms RANSAC50, SC2PCR, MAC, and PCR-99 by 465, 187, 9041, and 221 times, respectively. On the courtyard, it is 40, 198, 77, 8311, and 145 times faster than RANSAC10K, RANSAC50K, SC2PCR, MAC, and PCR-99, respectively. As MAC constructs a high-order graph and searches for the maximum clique, its time consumption significantly escalates with higher correspondence numbers and inlier ratios. In contrast, TCF shows minimal impact from correspondence quantity. It merely requires an additional 31 ms and 69 ms for trees and courtyard than the office, respectively, which underscores our high efficiency.

\subsection{Results on KITTI}
\begin{figure}[t]
	\centering
	\includegraphics[width = 8cm]{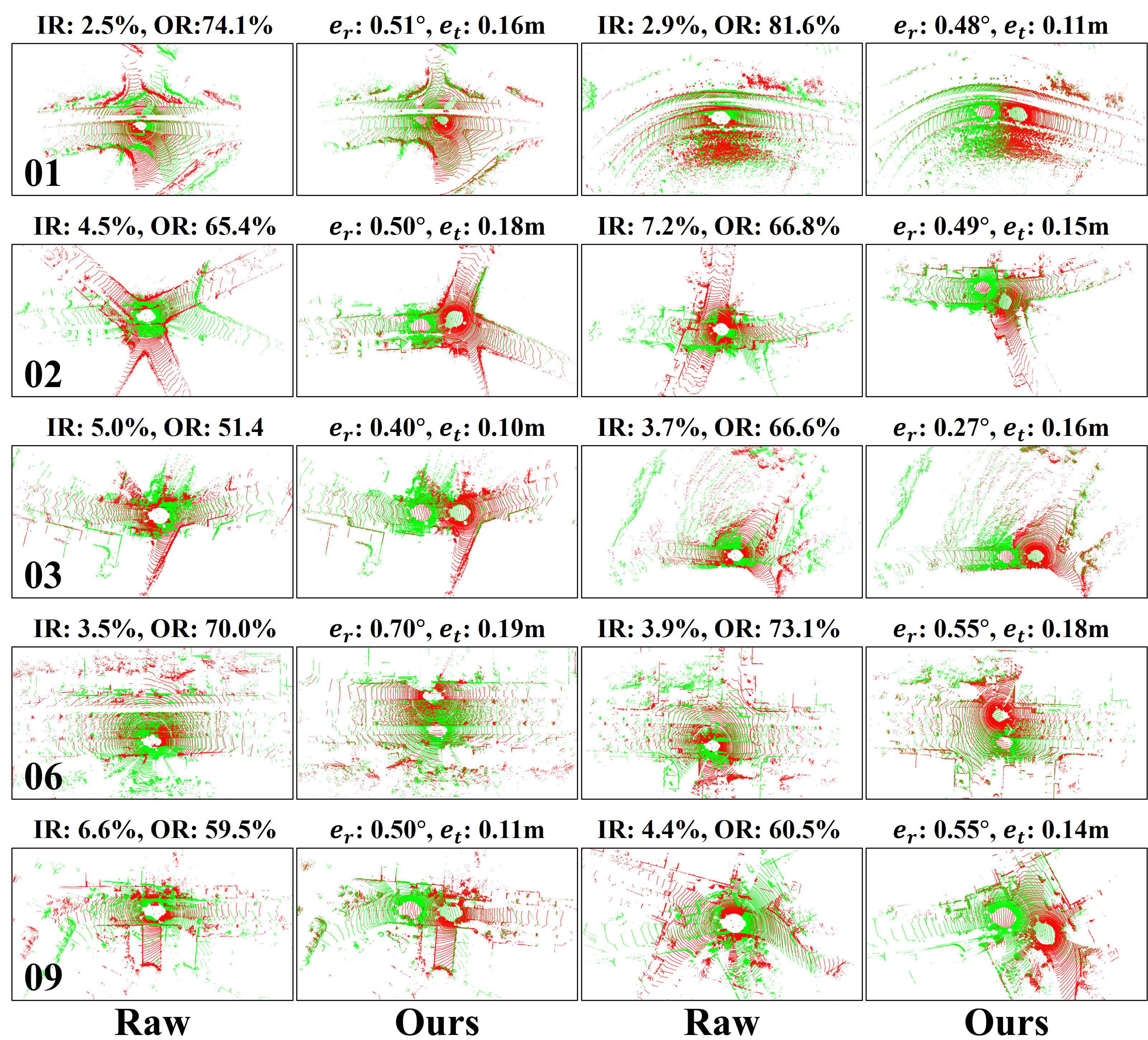}
	\caption{Visualization of registration results on the KITTI dataset. Red and green denote the source and target point clouds, respectively. IR refers to the inlier ratio, and OR denotes the overlap ratio. $e_r$ refers to the rotation error, and $e_t$ denotes the translation error.}\label{fig:kitti-registration}
\end{figure}

\textbf{Recall and Accuracy.} Table \ref{tab:eva-kitti-top-3} summarizes the registration results on KITTI dataset. Fig. \ref{fig:kitti-registration} depicts our registration results. The large offset (about 17 m) between point clouds degrades performance for all methods. Sequence 01 is a high-speed scenario characterized by significant distortion and feature deterioration, resulting in a low inlier ratio of 2.4\%. Despite this, our method achieves the highest recall (65\%). It attains the second-highest translation accuracy across all five sequences. Our rotation accuracy slightly lags behind SC2PCR and MAC but stays within a maximum deviation of 0.18\degree. These results underscore that our method excels in challenging environments like highways, exhibiting higher registration recall. 

\textbf{Running Efficiency.} Our method still demonstrates superior efficiency. In sequence 01, it outperforms RANSAC10K, RANSAC50K, TEASER, SC2PCR, MAC, and PCR-99 by 28, 141, 2, 23, 451, and 155 times, respectively. Similarly, in sequence 06, our speeds surpass RANSAC10K, RANSAC50K, TEASER, SC2PCR, MAC, and PCR-99 by 42, 205, 3, 35, 465,  2132, and 194 times, respectively. In sequence 09, our method executes in 54 ms, significantly outperforming RANSAC10K, RANSAC50K, TEASER, SC2PCR, MAC, and PCR-99, being 46, 221, 2, 28, 580, and 68 times faster, respectively. Our approach offers comparable or superior performance to baseline methods while significantly enhancing speed. The results demonstrate our effectiveness and efficiency with low-overlap data. Furthermore, these results reaffirm that our method elevates the RANSAC family to SOTA performance.
\subsection{Boosting Performance with TCF}
As illustrated in Table \ref{tab:boost-with-tcf}, we feed our consensus set into other method. The total runtime in the table comprises ours and the baseline methods. Our approach boosts trees' inlier ratio from 1.2\% to 98\%. All pipelines achieve 100\% recall, with deviations limited to 0.27° and 0.05 m. TEASER and PCR-99 enhance the recall by 30\% and 50\%, respectively. TEASER, SC2PCR, MAC, and PCR-99 are 332 ms, 9 s, 330 s, and 10 s faster than their original methods, respectively. We boost the inlier ratio in sequence 01 jumps from 2.4\% to 48\%. TEASER, SC2PCR and MAC demonstrate 12\%, 15\%, and 10\% recall improvement, respectively. TEASER, SC2PCR, MAC, and PCR-99 become faster by approximately 70 ms, 2 s, 30 s, and 12 s, respectively. The results prove that cascading our method with other methods effectively boosts registration performance and facilitates a qualitative leap in speed.
\begin{table}[!t]
	\renewcommand{\arraystretch}{1}
	\centering
	\caption{Performance improvement with our method. The runtime includes our method and baseline methods. $\mathcal{C}_1$ and $\rho_1$ mean raw correspondences and inlier ratios, respectively, while $\mathcal{C}_2$ and $\rho_2$ refer to those generated by our method.}
	\begin{tabular}{m{0.9cm}<{\centering}|m{0.3cm}<{\centering}m{0.4cm}<{\centering}m{0.4cm}<{\centering}m{0.9cm}<{\centering}|m{0.3cm}<{\centering}m{0.4cm}<{\centering}m{0.4cm}<{\centering}m{0.9cm}<{\centering}}
		\hline
		\multirow{5}{*}{Methods}&\multicolumn{4}{c|}{Trees}&\multicolumn{4}{c}{01}\\\cline{2-9}
		&\multicolumn{4}{c|}{$|\mathcal{C}_1| = 8242, \rho_1 = 1.2\%$}&\multicolumn{4}{c}{$|\mathcal{C}_1| = 3343, \rho_1 = 2.4\%$}\\
		&\multicolumn{4}{c|}{$|\mathcal{C}_2| = 60, \rho_2 = 98\%$}&\multicolumn{4}{c}{$|\mathcal{C}_2| = 121, \rho_2 = 48\%$}\\\cline{2-9}
		&$rr$&$e_r$&$e_t$&time&$rr$&$e_r$&$e_t$&time\\
		&(\%)&(\degree)&(cm)&(ms)&(\%)&(\degree)&(cm)&(ms)\\\hline
		\multirow{2}{*}{\makecell[c]{Ours+\\TEASER}}&100&0.24&0.05&59&64&0.54&0.25&96\\
		&$\uparrow$30&$\downarrow$0.11&$\downarrow$0.01&$\downarrow$332&$\uparrow$12&$\uparrow$0.07&$\downarrow$0.09&$\downarrow$70\\\cdashline{1-9} 
		\multirow{2}{*}{\makecell[c]{Ours+\\SC2PCR}}&100&0.18&0.02&63&64&0.50&0.24&87\\
		&$\uparrow$10&$\uparrow$0.02&$\downarrow$0.01&$\downarrow$8636&$\uparrow$15&$\uparrow$0.14&$\uparrow$0.03&$\downarrow$1714\\\cdashline{1-9} 
		\multirow{2}{*}{\makecell[c]{Ours+\\MAC}}&100&0.22&0.04&1672&64&0.51&0.24&5203\\
        &$-$&$\downarrow$0.03&$\downarrow$0.03&$\downarrow$329915&$\uparrow$10&$\uparrow$0.11&$\downarrow$0.02&$\downarrow$30,429\\\cdashline{1-9} 
		\multirow{2}{*}{\makecell[c]{Ours+\\PCR-99}}&100&0.27&0.04&59&63&0.94&0.33&121\\
		&$\uparrow$50&$\uparrow$0.03&$\downarrow$0.02&$\downarrow$10250&$\uparrow$8&$\downarrow$0.43&$\downarrow$0.12&$\downarrow$12098\\\hline
	\end{tabular}\label{tab:boost-with-tcf}
\end{table}
\subsection{Noise Study}
\begin{figure}[!t]
	\centering
	\subfigure[Recall]{\includegraphics[width=4.3cm] {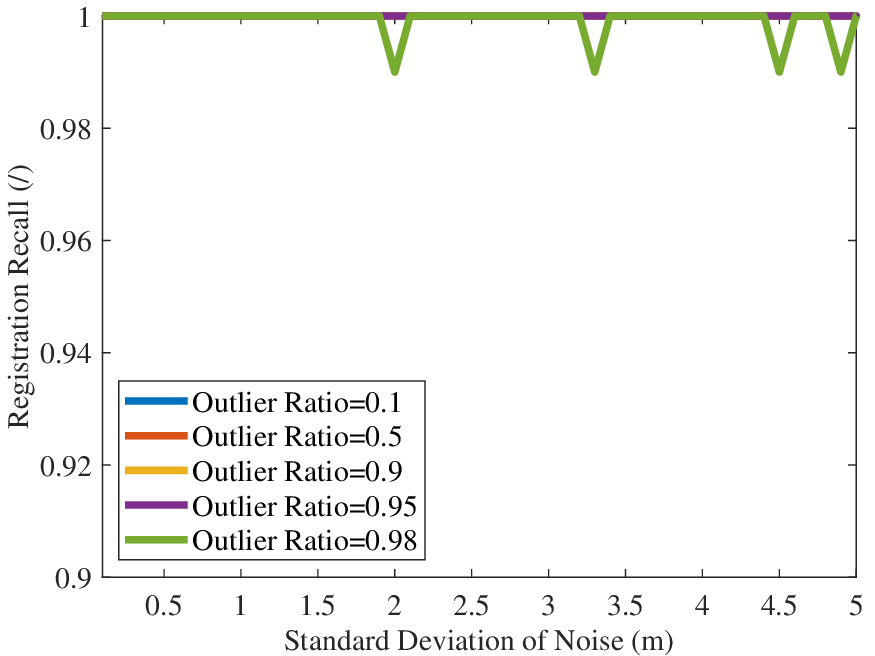}}
	\subfigure[Rotation and translation error]{\includegraphics[width=4.3cm] {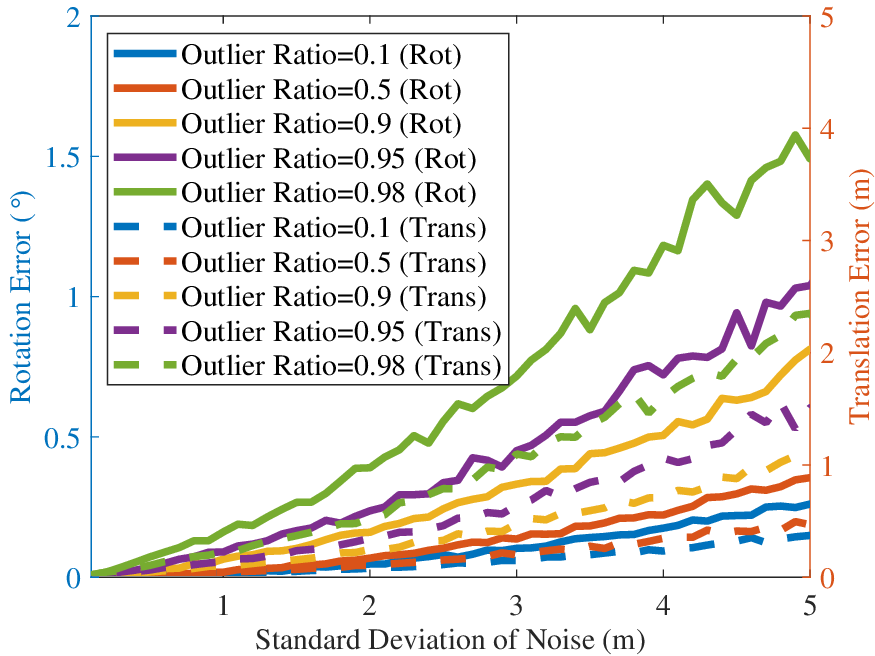}}
	\caption{Registration across various noise levels. (b): Rotation (left) and translation (right) errors are shown in one figure.}\label{fig:noise-registration}
\end{figure}

As shown in Fig. \ref{fig:noise-registration}, we evaluate the registration performance under various noise levels. In each, we randomly generate 3000 correspondences between -100 m and 100 m, satisfying the same transformation. Then, we replace 10\%, 50\%, 90\%, 95\%, and 98\% correspondences with random values between -100 m and 100 m as outliers. Finally, we add 50 levels of Gaussian noise, with standard deviations $\sigma_{noise}$ from 0.1 m to 5 m, to the remaining inliers. We run 100 tests to compute recall, average rotation error, and translation error. Registration is successful if the RMSE of true inliers is less than three times $\sigma_{noise}$. The results show an overall recall of over 98\%. With 98\% outliers, the translation error stays below the noise level, and the rotation error remains under 2\degree, highlighting the method's robustness.
\subsection{Ablation Study}
\textbf{Outlier Removal.} Table \ref{tab:ablation-correspondence} examines the contributions of our three RANSAC modules in outlier removal. IR is the inlier ratio while IN means the inlier number. In the arch, the slowest three-point RANSAC (3R) requires 3234 ms and achieves an inlier ratio of 27\%. The fastest one-point RANSAC (1R) completes in 20 ms, delivering an inlier rate of 21\%. Compared to 3R, 1R+2R+3R increases the inlier ratio by 58\% and reduces runtime by 3202 ms. In sequence 01, 1R remains the fastest, taking 29 ms with 8\% inliers. The slowest 3R requires 3139 ms and achieves 28\% inliers. 1R+2R+3R attains the highest inlier ratio at 38\%, reduces runtime by 3089 ms compared to 3R, and improves the inlier ratio by 30\% relative to 1R. The results illustrate that our method decreases iterations for 3R and preserves a high inlier ratio. 

\textbf{PCR Performance.} Table \ref{tab:ablation-PCR} examines the contributions of 3R and IRLS to final registration performance. In Arch, 3R+IRLS reduces rotation errors by 0.06° and translation errors by 0.02 m compared to using only 3R. It shows improvements of 0.04° and 0.01 m upon IRLS. It takes only 1-2 ms longer than 3R and IRLS. In sequence 06, 3R+IRLS enhances recall by 1\% over other pipelines. Compared to 3R, it reduces angle error by 0.35° and translation error by 0.07 m. Compared to IRLS, it reduces errors by 0.12° and 0.02 m. It takes just an extra 4 ms than the fastest 3R. These results indicate: (1) with most outliers removed by 1R+2R, the subsequent registration converges quickly, and (2) our overall pipeline (1R+2R+3R+IRLS) balances accuracy and efficiency.
\begin{table}[!t]
	\centering
	\caption{Outlier removal analysis for three RANSAC modules. 1R, 2R, and 3R denote one-point, two-point, and three-point RANSAC, respectively. IR is the inlier ratio. IN means the inlier number.}
	\begin{tabular}{m{1.4cm}<{\centering}|m{0.6cm}<{\centering}m{0.6cm}<{\centering}m{0.6cm}<{\centering}|m{0.6cm}<{\centering}m{0.6cm}<{\centering}m{0.6cm}<{\centering}}  
		\hline
		\multirow{3}{*}{\makecell[c]{RANSAC \\Module}}&\multicolumn{3}{c|}{Arch}&\multicolumn{3}{c}{01}\\\cline{2-7}
		&IR (\%)&IN (/)&Time (ms)&IR (\%)&IN (/)&Time (ms)\\\hline
		1R&21&\textbf{76}&\textbf{20}&8&\textbf{57}&\textbf{29}\\
		2R&26&48&672&12&27&115\\
		3R&27&16&3234&28&24&3139\\
		1R+2R&30&35&\underline{22}&19&21&\underline{34}\\
		1R+3R&\underline{63}&\underline{57}&25&35&\underline{47}&43\\
		2R+3R&62&35&680&\underline{36}&25&118\\
		1R+2R+3R&\textbf{85}&20&32&\textbf{38}&16&50\\\hline
	\end{tabular}\label{tab:ablation-correspondence}
\end{table}
\begin{table}[!t]
	\renewcommand{\arraystretch}{1}
	\centering
	\caption{Ablation experiments on registration performance for 3R and IRLS. Both 3R and IRLS can compute a 6-DOF pose.}
	\begin{tabular}{c|m{0.22cm}<{\centering}m{0.46cm}<{\centering}|m{0.32cm}<{\centering}m{0.3cm}<{\centering}m{0.3cm}<{\centering}m{0.38cm}<{\centering}|m{0.32cm}<{\centering}m{0.3cm}<{\centering}m{0.3cm}<{\centering}m{0.38cm}<{\centering}}  
		\hline
		\multirow{3}{*}{}&	\multirow{3}{*}{3R}&\multirow{3}{*}{IRLS}&\multicolumn{4}{c|}{Arch}&\multicolumn{4}{c}{06}\\\cline{4-11}
		&&&$rr$ (\%)&$e_r$ (\degree)&$e_t$ (m)&time (ms)&$rr$ (\%)&$e_r$ (\degree)&$e_t$ (m)&time (ms)\\\hline
        1R+2R&\checkmark&&\textbf{88}&0.19&0.06&\textbf{30}&\underline{74}&0.86&0.30&\textbf{39}\\
        1R+2R&&\checkmark&\textbf{88}&\underline{0.17}&\underline{0.05}&\underline{31}&\underline{74}&\underline{0.63}&\underline{0.25}&\underline{40}\\
        1R+2R&\checkmark&\checkmark&\textbf{88}&\textbf{0.13}&\textbf{0.04}&32&\textbf{75}&\textbf{0.51}&\textbf{0.23}&43\\\hline
	\end{tabular}\label{tab:ablation-PCR}
\end{table}
\section{Conclusion}\label{sec:conclusion}
We propose a two-stage consensus filtering method for correspondence-based registration. Our approach restores the RANSAC family to SOTA performance. By combining length consistency with angle consistency, we eliminate the majority of outliers, thus reducing the requisite iterations by three-point RANSAC. We combine three RASNAC modules and iterative weighted least squares into a complete registration pipeline that accelerates registration without compromising accuracy. Experimental results demonstrate a three-orders-of-magnitude speed enhancement while maintaining high registration quality. Moreover, integrating our method with existing approaches can notably improve efficiency and recall.
\bibliographystyle{IEEEtran}
\bibliography{refs}

\begin{thebibliography}{10}
\providecommand{\url}[1]{#1}
\csname url@samestyle\endcsname
\providecommand{\newblock}{\relax}
\providecommand{\bibinfo}[2]{#2}
\providecommand{\BIBentrySTDinterwordspacing}{\spaceskip=0pt\relax}
\providecommand{\BIBentryALTinterwordstretchfactor}{4}
\providecommand{\BIBentryALTinterwordspacing}{\spaceskip=\fontdimen2\font plus
\BIBentryALTinterwordstretchfactor\fontdimen3\font minus
  \fontdimen4\font\relax}
\providecommand{\BIBforeignlanguage}[2]{{%
\expandafter\ifx\csname l@#1\endcsname\relax
\typeout{** WARNING: IEEEtran.bst: No hyphenation pattern has been}%
\typeout{** loaded for the language `#1'. Using the pattern for}%
\typeout{** the default language instead.}%
\else
\language=\csname l@#1\endcsname
\fi
#2}}
\providecommand{\BIBdecl}{\relax}
\BIBdecl

\bibitem{2022-CVPR-SC2PCR}
Z.~Chen, K.~Sun, F.~Yang, and W.~Tao, ``Sc2-pcr: A second order spatial
  compatibility for efficient and robust point cloud registration,'' in
  \emph{2022 IEEE/CVF Conference on Computer Vision and Pattern Recognition
  (CVPR)}, 2022, pp. 13\,211--13\,221.

\bibitem{2023-CVPR-MAC}
X.~Zhang, J.~Yang, S.~Zhang, and Y.~Zhang, ``3d registration with maximal
  cliques,'' in \emph{2023 IEEE/CVF Conference on Computer Vision and Pattern
  Recognition (CVPR)}, 2023, pp. 17\,745--17\,754.

\bibitem{2017-CVPR-3DMatch}
A.~Zeng, S.~Song, M.~Nießner, M.~Fisher, J.~Xiao, and T.~Funkhouser,
  ``3dmatch: Learning local geometric descriptors from rgb-d reconstructions,''
  in \emph{2017 IEEE Conference on Computer Vision and Pattern Recognition
  (CVPR)}, 2017, pp. 199--208.

\bibitem{2016-ECCV-FGR}
Q.-Y. Zhou, J.~Park, and V.~Koltun, ``Fast global registration,'' in
  \emph{Computer Vision--ECCV 2016: 14th European Conference, Amsterdam, The
  Netherlands, October 11-14, 2016, Proceedings, Part II 14}.\hskip 1em plus
  0.5em minus 0.4em\relax Springer, 2016, pp. 766--782.

\bibitem{2024-RAL-CCAG}
Y.~Wang, P.~Zhou, G.~Geng, L.~An, and Y.~Liu, ``Ccag: End-to-end point cloud
  registration,'' \emph{IEEE Robotics and Automation Letters}, vol.~9, no.~1,
  pp. 435--442, 2024.

\bibitem{2020-IJGI-ANIS}
P.~Shi, Q.~Ye, and L.~Zeng, ``A novel indoor structure extraction based on
  dense point cloud,'' \emph{ISPRS International Journal of Geo-Information},
  vol.~9, no.~11, p. 660, 2020.

\bibitem{2023-JAG-100FPS}
P.~Shi, J.~Li, and Y.~Zhang, ``Lidar localization at 100 fps: A map-aided and
  template descriptor-based global method,'' \emph{International Journal of
  Applied Earth Observation and Geoinformation}, vol. 120, p. 103336, 2023.

\bibitem{2020-Sensors-ANLC}
Q.~Ye, P.~Shi, K.~Xu, P.~Gui, and S.~Zhang, ``A novel loop closure detection
  approach using simplified structure for low-cost lidar,'' \emph{Sensors},
  vol.~20, no.~8, p. 2299, 2020.

\bibitem{2023-ISPRSJ-OPD}
P.~Shi, J.~Li, and Y.~Zhang, ``A fast lidar place recognition and localization
  method by fusing local and global search,'' \emph{ISPRS Journal of
  Photogrammetry and Remote Sensing}, vol. 202, pp. 637--651, 2023.

\bibitem{2024-TIV-ANHE}
P.~Shi, Y.~Xiao, W.~Chen, J.~Li, and Y.~Zhang, ``A new horizon: Employing map
  clustering similarity for lidar-based place recognition,'' \emph{IEEE
  Transactions on Intelligent Vehicles}, 2024.

\bibitem{2022-CVPR-Regtr}
Z.~J. Yew and G.~H. Lee, ``Regtr: End-to-end point cloud correspondences with
  transformers,'' in \emph{Proceedings of the IEEE/CVF conference on computer
  vision and pattern recognition}, 2022, pp. 6677--6686.

\bibitem{2024-TPR-Survey-SLAM}
Y.~Zhang, P.~Shi, and J.~Li, ``3d lidar slam: A survey,'' \emph{The
  Photogrammetric Record}, 2024.

\bibitem{2023-Survey-arXiv-LPRF}
P.~Shi, Y.~Zhang, and J.~Li, ``Lidar-based place recognition for autonomous
  driving: A survey,'' \emph{arXiv preprint arXiv:2306.10561}, 2023.

\bibitem{1992-TPAMI-ICP}
P.~Besl and N.~D. McKay, ``A method for registration of 3-d shapes,''
  \emph{IEEE Transactions on Pattern Analysis and Machine Intelligence},
  vol.~14, no.~2, pp. 239--256, 1992.

\bibitem{2021-AGCS-LIFM}
P.~Shi, Q.~Ye, Z.~Shaoming, and D.~Haifeng, ``Localization initialization for
  multi-beam lidar considering indoor scene feature,'' \emph{Acta Geodaetica et
  Cartographica Sinica}, vol.~50, pp. 1594--1604, 2021.

\bibitem{shi2024indoor}
P.~SHI, J.~LI, X.~LIU, and Y.~ZHANG, ``Indoor cylinders guided lidar global
  localization and loop closure detection,'' \emph{Geomatics and Information
  Science of Wuhan University}, vol.~49, no.~7, pp. 1088--1099, 2024.

\bibitem{2021-TRO-Teaser}
H.~Yang, J.~Shi, and L.~Carlone, ``Teaser: Fast and certifiable point cloud
  registration,'' \emph{IEEE Transactions on Robotics}, vol.~37, no.~2, pp.
  314--333, 2021.

\bibitem{2020-TGRS-CG-SAC}
S.~Quan and J.~Yang, ``Compatibility-guided sampling consensus for 3-d point
  cloud registration,'' \emph{IEEE Transactions on Geoscience and Remote
  Sensing}, vol.~58, no.~10, pp. 7380--7392, 2020.

\bibitem{2019-ICCV-DCP}
Y.~Wang and J.~M. Solomon, ``Deep closest point: Learning representations for
  point cloud registration,'' in \emph{Proceedings of the IEEE/CVF
  international conference on computer vision}, 2019, pp. 3523--3532.

\bibitem{2023-TPAMI-QGORE}
J.~Li, P.~Shi, Q.~Hu, and Y.~Zhang, ``Qgore: Quadratic-time guaranteed outlier
  removal for point cloud registration,'' \emph{IEEE Transactions on Pattern
  Analysis and Machine Intelligence}, vol.~45, no.~9, pp. 11\,136--11\,151,
  2023.

\bibitem{1981-CACM-RANSAC}
M.~A. Fischler and R.~C. Bolles, ``Random sample consensus: a paradigm for
  model fitting with applications to image analysis and automated
  cartography,'' \emph{Communications of the ACM}, vol.~24, no.~6, pp.
  381--395, 1981.

\bibitem{2009-ICRA-FPFH}
R.~B. Rusu, N.~Blodow, and M.~Beetz, ``Fast point feature histograms (fpfh) for
  3d registration,'' in \emph{2009 IEEE international conference on robotics
  and automation}.\hskip 1em plus 0.5em minus 0.4em\relax IEEE, 2009, pp.
  3212--3217.

\bibitem{2018-CVPR-GC-RANSAC}
D.~Barath and J.~Matas, ``Graph-cut ransac,'' in \emph{2018 IEEE/CVF Conference
  on Computer Vision and Pattern Recognition}, 2018, pp. 6733--6741.

\bibitem{2024-arXiv-PCR-99}
S.~H. Lee, J.~Civera, and P.~Vandewalle, ``Pcr-99: A practical method for point
  cloud registration with 99\% outliers,'' \emph{arXiv preprint
  arXiv:2402.16598}, 2024.

\bibitem{lusk2021clipper}
P.~C. Lusk, K.~Fathian, and J.~P. How, ``Clipper: A graph-theoretic framework
  for robust data association,'' in \emph{2021 IEEE International Conference on
  Robotics and Automation (ICRA)}.\hskip 1em plus 0.5em minus 0.4em\relax IEEE,
  2021, pp. 13\,828--13\,834.

\bibitem{shi2021robin}
J.~Shi, H.~Yang, and L.~Carlone, ``Robin: a graph-theoretic approach to reject
  outliers in robust estimation using invariants,'' in \emph{2021 IEEE
  International Conference on Robotics and Automation (ICRA)}.\hskip 1em plus
  0.5em minus 0.4em\relax IEEE, 2021, pp. 13\,820--13\,827.

\bibitem{1992-IVC-Point-Plane-ICP}
Y.~Chen and G.~Medioni, ``Object modelling by registration of multiple range
  images,'' \emph{Image and vision computing}, vol.~10, no.~3, pp. 145--155,
  1992.

\bibitem{2019-RSS-GICP}
A.~Segal, D.~Haehnel, and S.~Thrun, ``Generalized-icp.'' in \emph{Robotics:
  science and systems}, vol.~2, no.~4.\hskip 1em plus 0.5em minus 0.4em\relax
  Seattle, WA, 2009, p. 435.

\bibitem{2016-TPAMI-GOICP}
J.~Yang, H.~Li, D.~Campbell, and Y.~Jia, ``Go-icp: A globally optimal solution
  to 3d icp point-set registration,'' \emph{IEEE Transactions on Pattern
  Analysis and Machine Intelligence}, vol.~38, no.~11, pp. 2241--2254, 2016.

\bibitem{yang2020graduated}
H.~Yang, P.~Antonante, V.~Tzoumas, and L.~Carlone, ``Graduated non-convexity
  for robust spatial perception: From non-minimal solvers to global outlier
  rejection,'' \emph{IEEE Robotics and Automation Letters}, vol.~5, no.~2, pp.
  1127--1134, 2020.

\bibitem{2019-ATG-DGCNN}
Y.~Wang, Y.~Sun, Z.~Liu, S.~E. Sarma, M.~M. Bronstein, and J.~M. Solomon,
  ``Dynamic graph cnn for learning on point clouds,'' \emph{ACM Transactions on
  Graphics (tog)}, vol.~38, no.~5, pp. 1--12, 2019.

\bibitem{2015-Nips-Pointer-Network}
O.~Vinyals, M.~Fortunato, and N.~Jaitly, ``Pointer networks,'' \emph{Advances
  in neural information processing systems}, vol.~28, 2015.

\bibitem{2019-ICCV-FCGF}
C.~Choy, J.~Park, and V.~Koltun, ``Fully convolutional geometric features,'' in
  \emph{Proceedings of the IEEE/CVF international conference on computer
  vision}, 2019, pp. 8958--8966.

\bibitem{2018-CVPR-PPFNET}
H.~Deng, T.~Birdal, and S.~Ilic, ``Ppfnet: Global context aware local features
  for robust 3d point matching,'' in \emph{Proceedings of the IEEE conference
  on computer vision and pattern recognition}, 2018, pp. 195--205.

\bibitem{2018-ECCV-Ppf-foldnet}
------, ``Ppf-foldnet: Unsupervised learning of rotation invariant 3d local
  descriptors,'' in \emph{Proceedings of the European conference on computer
  vision (ECCV)}, 2018, pp. 602--618.

\bibitem{2021-TPAMI-LEAD}
M.~Marcon, R.~Spezialetti, S.~Salti, L.~Silva, and L.~Di~Stefano,
  ``Unsupervised learning of local equivariant descriptors for point clouds,''
  \emph{IEEE Transactions on Pattern Analysis and Machine Intelligence},
  vol.~44, no.~12, pp. 9687--9702, 2021.

\bibitem{2023-AGCS-SACR}
L.~Jiayuan, Z.~Yongjun, A.~Mingyao, and H.~Qingwu, ``Scale-adaptive cauchy
  robust estimation based on progressive optimization and its applications,''
  \emph{Acta Geodaetica et Cartographica Sinica}, vol.~52, no.~1, pp. 61--70,
  2023.

\bibitem{2015-ISPRSJ-GCRO}
P.~W. Theiler, J.~D. Wegner, and K.~Schindler, ``Globally consistent
  registration of terrestrial laser scans via graph optimization,'' \emph{ISPRS
  journal of photogrammetry and remote sensing}, vol. 109, pp. 126--138, 2015.

\bibitem{2012-Dataset-CVPR-KITTI}
A.~Geiger, P.~Lenz, and R.~Urtasun, ``Are we ready for autonomous driving? the
  kitti vision benchmark suite,'' in \emph{2012 IEEE Conference on Computer
  Vision and Pattern Recognition}, 2012, pp. 3354--3361.

\bibitem{2009-ICCVW-ISS}
Y.~Zhong, ``Intrinsic shape signatures: A shape descriptor for 3d object
  recognition,'' in \emph{2009 IEEE 12th international conference on computer
  vision workshops, ICCV Workshops}.\hskip 1em plus 0.5em minus 0.4em\relax
  IEEE, 2009, pp. 689--696.

\end{thebibliography}
\end{document}